%% file: main.tex
\documentclass[10pt,twocolumn,letterpaper]{article}

\usepackage{cvpr}              
\usepackage[accsupp]{axessibility}
\usepackage{graphicx}
\usepackage{amsmath}
\usepackage{amssymb}
\usepackage{booktabs}
\usepackage{color}
\usepackage{xcolor}
\usepackage{ctable}
\usepackage{colortbl}
\usepackage{lipsum}
\usepackage{multirow}
\usepackage{soul}
\usepackage{xspace}\xspace
\usepackage{adjustbox}
\usepackage{array}
\usepackage{epsfig}
\usepackage{bbding}
\usepackage{dblfloatfix}
\usepackage{transparent}
\usepackage{diagbox}
\usepackage{listings}
\usepackage{nicefrac} 
\usepackage{times}
\usepackage{url}
\usepackage[utf8]{inputenc} 
\usepackage[T1]{fontenc}    
\usepackage{amsfonts}       
\usepackage{microtype}      
\usepackage{bm}
\usepackage{bbm}
\usepackage{amsbsy}
\usepackage{cases}
\usepackage{enumitem}
\usepackage{subcaption}
\usepackage[titletoc]{appendix}
\usepackage{grffile}
\usepackage{diagbox}
\usepackage{pifont}
\usepackage{multicol}
\usepackage{subcaption}
\usepackage{comment}
\definecolor{citecolor}{HTML}{0071bc}
\usepackage[pagebackref,breaklinks,colorlinks]{hyperref}
\definecolor{Gray}{gray}{0.92}
\definecolor{darkgreen}{rgb}{0.13, 0.55, 0.13}
\definecolor{Highlight}{HTML}{39b54a} 

\usepackage[capitalize]{cleveref}
\crefname{section}{Sec.}{Secs.}
\Crefname{section}{Section}{Sections}
\Crefname{table}{Table}{Tables}
\crefname{table}{Tab.}{Tabs.}

\def\cvprPaperID{6678} 
\def\confName{CVPR}
\def\confYear{2023}

\newcommand{\ourMethod}{CAPE}
\newcommand{\highNDS}{$61.0\%$}
\newcommand{\highmAP}{$52.5\%$}

\begin{document}

\title{CAPE: Camera View Position Embedding for Multi-View 3D Object Detection}


\author{
Kaixin Xiong\thanks{Equal contribution. ~~\textsuperscript{\dag}Corresponding author. This work is done when Kaixin Xiong is an intern at Baidu Inc. } $  ^{,1}$, \ \  
Shi Gong$^{*,2}$, \ \  
Xiaoqing Ye$^{*,2}$, \ \  
Xiao Tan$^{2}$, \ \  
Ji Wan$^{2}$, \\
Errui Ding$^{2}$, \ \
Jingdong Wang$^{\dag, 2}$, \ \
Xiang Bai$^{1}$\\
\textsuperscript{1}Huazhong University of Science and Technology, 
\textsuperscript{2}Baidu Inc.\\
{\tt\small kaixinxiong@hust.edu.cn,} \ \
{\tt\small \{gongshi, yexiaoqing\}@baidu.com} \ \
{\tt\small wangjingdong@outlook.com} \ \
}
\maketitle

\begin{abstract}

In this paper, we address the problem of detecting 3D objects from multi-view images.
Current query-based methods rely on global 3D position embeddings (PE) to learn the geometric correspondence between images and 3D space.
We claim that directly interacting 2D image features with global 3D PE 
could increase the difficulty of learning view transformation due to the variation of camera extrinsics.
Thus we propose a novel method based on \textbf{CA}mera view \textbf{P}osition \textbf{E}mbedding, called CAPE.
We form the 3D position embeddings under the local camera-view coordinate system instead of the global coordinate system, such that 3D position embedding is free of encoding camera extrinsic parameters. 
Furthermore, we extend our CAPE to temporal modeling by exploiting the object queries of previous frames and encoding the ego motion for boosting $3$D object detection. CAPE achieves the state-of-the-art performance (\highNDS{} NDS and \highmAP{} mAP) among all LiDAR-free methods on nuScenes dataset.
Codes and models are available.\footnote{Codes of \href{https://github.com/PaddlePaddle/Paddle3D}{Paddle3D} and \href{https://github.com/kaixinbear/CAPE}{PyTorch Implementation}.}
\end{abstract}

\input{sections/1_intro}

\input{sections/2_relatedwork}
\input{sections/3_method}
\input{sections/4_experiments}
\input{sections/5_conclusion}

{\small
\bibliographystyle{ieee_fullname}
\bibliography{egbib}
}


\end{document}

%% file: sections/1_intro.tex
\section{Introduction}
\label{sec:intro}
3D perception from multi-view cameras is a promising solution for autonomous driving due to its low cost and rich semantic knowledge. Given multiple sensors equipped on autonomous vehicles, how to perform end-to-end 3D perception integrating all features into a unified space is of critical importance. In contrast to traditional perspective-view perception that relies on post-processing to fuse the predictions from each monocular view~\cite{wang2021fcos3d, wang2022probabilistic} into the global 3D space, perception in the bird's-eye-view (BEV) is straightforward and thus arises increasing attention due to its unified representation for 3D location and scale, and easy adaptation for downstream tasks such as motion planning.

\begin{figure}[t]
\centering
\includegraphics[width=1\linewidth]{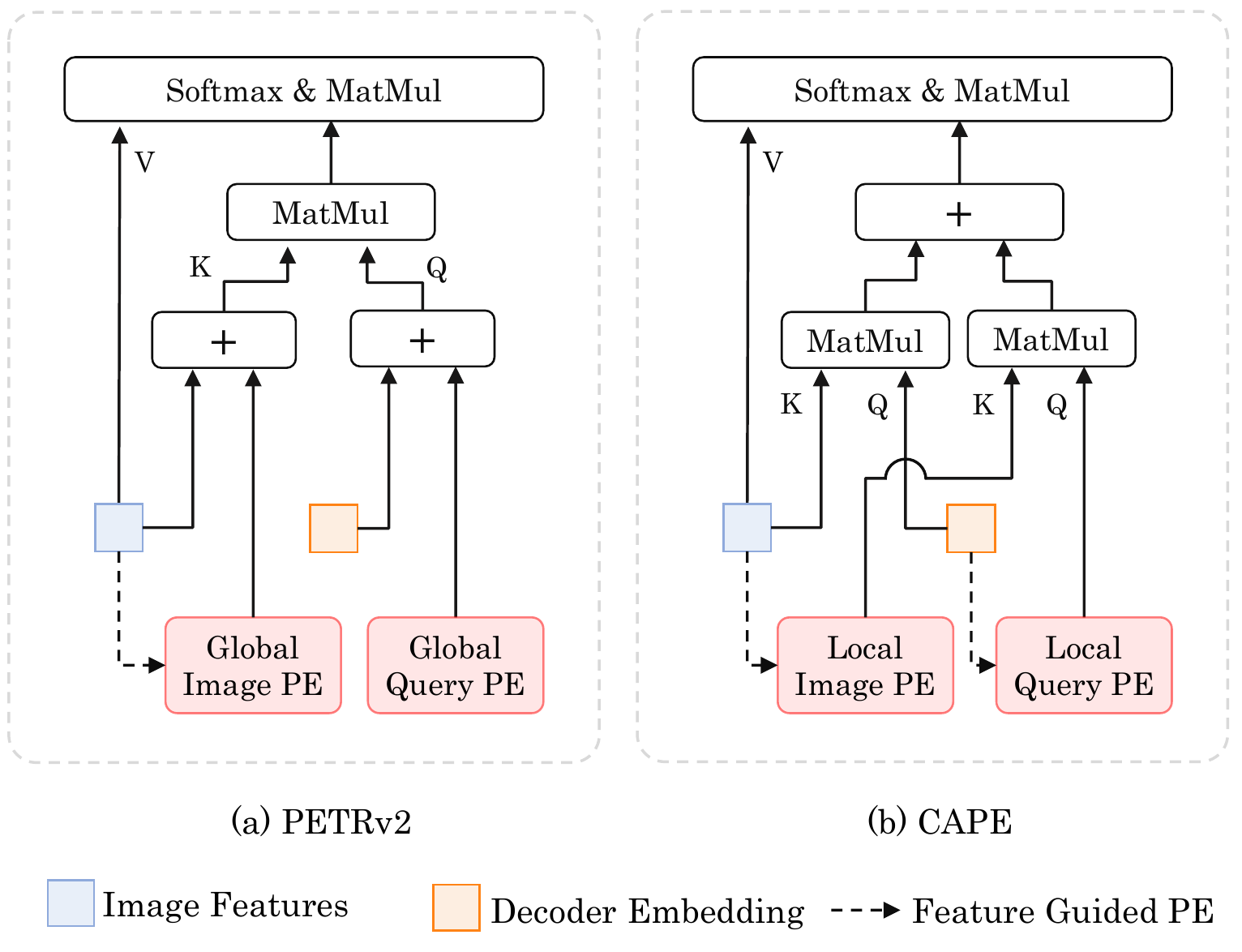}
\caption{
Comparison of the network structure between PETRv2 and our proposed \ourMethod{}. (a) In PETRv2, position embedding of queries and keys are in the global system. (b) In \ourMethod{}, position embeddings of queries and keys are within the local system of each view. Bilateral cross-attention is adopted to compute attention weights in the local and global systems independently.}
\centering
\label{fig:intro1}
\vspace{-10pt}
\end{figure}

The camera-based BEV perception is to predict 3D geometric outputs given the 2D features and thus the vital challenge is to learn the view transformation relationship between 2D and 3D space.
According to whether the explicit dense BEV representation is constructed, 
existing BEV approaches could be divided into two categories: the explicit BEV representation methods and the implicit BEV representation methods. 
The former constructs an explicit BEV feature map by lifting the 2D perspective-view features to 3D space~\cite{li2022bevformer,philion2020lift, huang2021bevdet}. 
The latter mainly follow DETR-based~\cite{carion2020end} approaches in an end-to-end manner. Without projection or lift operation, those methods \cite{zhou2022cross, liu2022petr,liu2022petrv2} implicitly encode the 3D global information into 3D position embedding (3D PE) to obtain 3D position-aware multi-view features, which is shown in Figure~\ref{fig:intro1}(a).

\begin{figure}[t]
\centering
\includegraphics[width=1\linewidth]{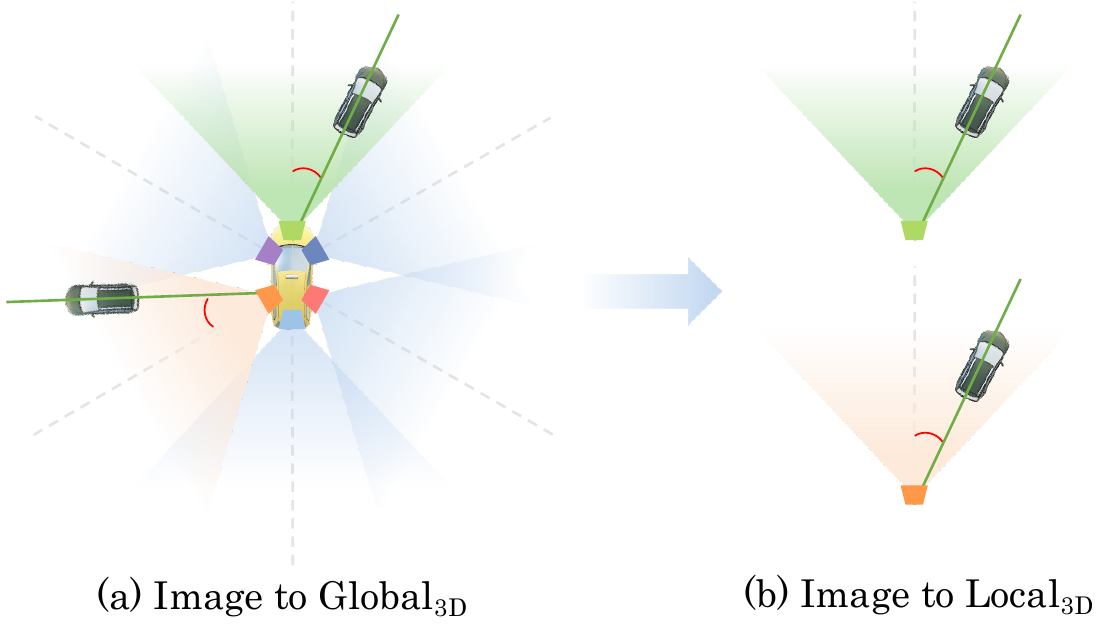}
\caption{View transformation comparison. In previous methods, view transformation is learned from the image to the global 3D coordinate system directly. In our method, the view transformation is learned from image to local (camera) coordinate system.}
\centering
\label{fig:intro2}
\vspace{-10pt}
\end{figure}

Though learning the transformation from 2D images to 3D global space is straightforward, we reveal that the interaction
in the global space for the query embeddings and 3D position-aware multi-view features hinders performance. The reasons are two-fold. For one thing, defining each camera coordinate system as the 3D local space, we find that the view transformation couples the 2D image-to-local transformation and the local-to-global transformation together. Thus the network is forced to differentiate variant camera extrinsics in the high-dimensional embedding space for 3D predictions in the global system, while the local-to-global relationship is a simple rigid transformation. For another, we believe the view-invariant transformation paradigm from 2D image to 3D local space is easier to learn, compared to directly transforming into 3D global space.
For example, though two vehicles in two views have similar appearances in image features, the network is forced to learn different view transformations, as depicted in Figure~\ref{fig:intro2}~(a).

To ease the difficulty in view transformation from 2D image to global space, we propose a simple yet effective approach based on local view position embedding, called \ourMethod{}, which performs 3D position embedding in the local system of each camera instead of the 3D global space. As depicted in Figure~\ref{fig:intro2} (b), our approach learns the view transformation from 2D image to local 3D space, which eliminates the variances of view transformation caused by different camera extrinsics. 




Specially,
as for key 3D PE, we transform camera frustum into 3D coordinates in the camera system using camera intrinsics only, then encoded by a simple MLP layer.
As for query 3D PE, we convert the 3D reference points defined in the global space into the local camera system with camera extrinsics only, then encoded by a simple MLP layer. 
Inspired by ~\cite{meng2021conditional, liu2022petrv2}, we obtain the 3D PE with the guidance of image features and decoder embeddings, for keys and queries, respectively.
Given that 3D PE is in the local space whereas the output queries are defined in the global coordinate system, 
we adopt the bilateral attention mechanism
to avoid the mixture of embeddings in different representation spaces, as shown in Figure~\ref{fig:intro1}(b).

We further extend \ourMethod{} to integrate multi-frame temporal information to boost the 3D object detection performance, named \ourMethod{}-T. Different from previous methods that either warp the explicit BEV features using ego-motion~\cite{li2022bevformer, 2022BEVDet4D} or encode the ego-motion into the position embedding~\cite{liu2022petrv2}, we adopt separated sets of object queries for each frame and encode the ego-motion to fuse the queries.

We summarize our key contributions as follows: 
\begin{itemize}
\vspace*{-1mm}
\item
We propose a novel multi-view 3D detection method, called \ourMethod{}, based on camera-view position embedding, which eliminates the variances of view transformation caused by different camera extrinsics.
\vspace*{-2mm}
\item
We further generalize our \ourMethod{} to temporal modeling, by exploiting the object queries of previous frames and leveraging the ego-motion explicitly for boosting 3D object detection and velocity estimation.

\vspace*{-2mm}
\item
Extensive experiments on the nuScenes dataset show the effectiveness of our proposed approach and we achieve the state-of-the-art among all LiDAR-free methods on the challenging nuScenes benchmark.
\vspace*{-1mm}
\end{itemize}


%% file: sections/2_relatedwork.tex
\begin{figure*}[t]
\centering
\includegraphics[width=1\linewidth]{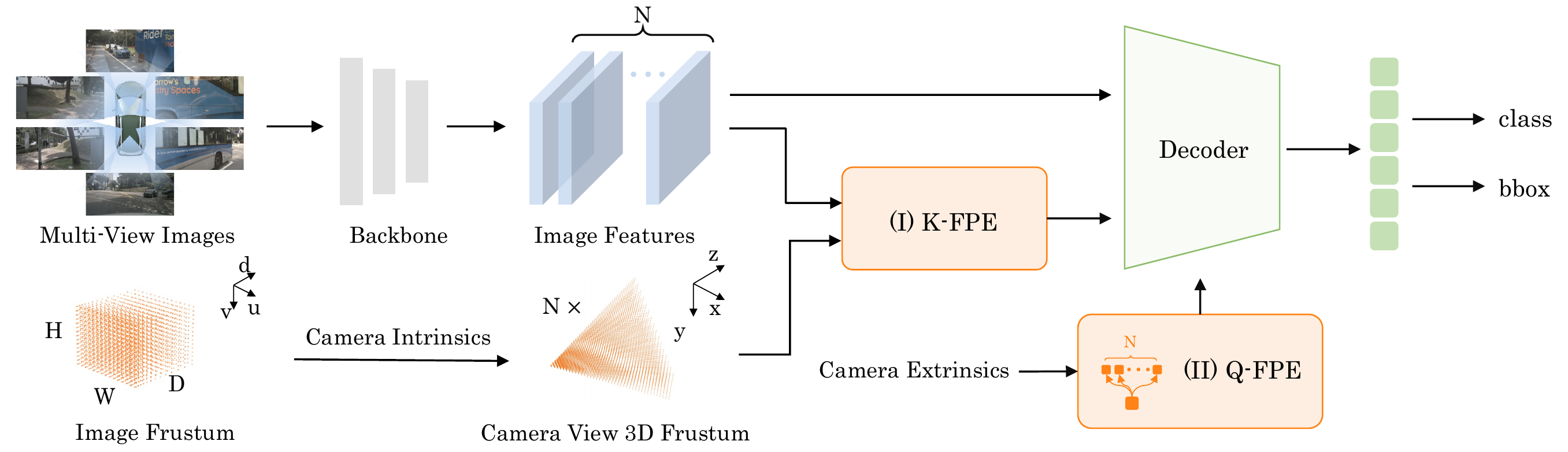}
\caption{The overview of \ourMethod{}. The multi-view images are fed into the backbone network to extract the 2D features including $N$ views. Key position embedding is formed by transforming camera-view frustum points to 3D coordinates in the camera system with camera intrinsics. Images features are used to guide the key position embedding in K-FPE; Query positional embedding is formed by converting the global 3D reference points into $N$ camera-view coordinates with camera extrinsics. Then we encode them under the guidance of decoder embeddings in Q-FPE. The decoder embeddings are updated via the interaction with image features in the decoder. The updated decoder embeddings are used to predict 3D bounding boxes and object classes.}
\vspace{-10pt}
\centering
\label{fig:overview}
\end{figure*}

\section{Related Work}
\subsection{DETR-based 2D Detection}
DETR ~\cite{carion2020end} is the pioneering work that successfully adopts transformers~\cite{vaswani2017attention} in the object detection task. It adopts a set of learnable queries to perform cross-attention and treat the matching process as a set prediction case. Many follow-up methods~\cite{zhu2020deformable, chen2022group, wang2022anchor, liu2022dab, li2022dn} focus on addressing the slow convergence problem in the training phase. 
For example, 
Conditional DETR~\cite{meng2021conditional} decouples the items in attention into spatial and content items, which eliminates the noises in cross attention and leads to fast convergence.

\subsection{Monocular 3D Detection}
Monocular 3D Detection task is highly related to multi-view 3D object detection since they both require restoring the depth information from images.  
The methods can be roughly grouped into two categories: pure image-based methods and depth-guided methods. Pure image-based methods mainly learn depth information from objects' apparent size and geometry constraints provided by eight keypoints or pin-hole model ~\cite{brazil2019m3d,mousavian20173d, li2020rtm3d,li2019gs3d,lian2021geometry,liu2022learning,wang2021depth,lu2021GUPNet}. 
Depth-guided methods need extra data sources such as point clouds and depth images in the training phase~\cite{CaDDN, ye2020monocular, ma2020rethinking, qian2020end, chong2021monodistill}. Pseudo-LiDAR~\cite{wang2019pseudo} converts pixels to pseudo point clouds and then feeds them into a LiDAR-based detector~\cite{shi2019pointrcnn, liu2020tanet, ye2020monocular,deng2021voxel}. DD3D~\cite{park2021pseudo} claims that pre-training paradigms could replace the pseudo-lidar paradigm. The quality of depth estimation would have a large influence on those methods.
   
\subsection{Multi-View 3D Detection}
Multi-view 3D detection aims to predict 3D bounding boxes in the global system from multi-cameras. 
Previous methods~\cite{wang2021fcos3d, wang2022probabilistic, epropnp} mostly extend from monocular 3D object detection. 
Those methods cannot leverage the geometric information in multi-view images. 
Recently, several methods~\cite{roddick2018orthographic, roddick2020predicting, philion2020lift, gong2022gitnet, li2022bevformer} attempt to percept objects in the global system using explicit bird's-eye view (BEV) maps. LSS~\cite{philion2020lift} conducts view transform via predicting depth distribution and lift images onto BEV.
BEVFormer\cite{li2022bevformer} exploits  spatial and temporal information through predefined grid-shaped BEV queries. 
BEVDepth~\cite{li2022bevdepth} leverages point clouds as the depth supervision and encodes camera parameters into the depth sub-network.

Some methods learn implicit BEV features following DETR~\cite{carion2020end} paradigm. These methods mainly initialize 3D sparse object queries and interact with 2D features by attention to directly perform 3D object detection. For example, DETR3D~\cite{wang2022detr3d} samples 2D features from the projected 3D reference points and then conducts local cross attention to update the queries. PETR~\cite{liu2022petr} proposes 3D position embedding in the global system and then conducts global cross attention to update the queries. 
PETRv2~\cite{liu2022petrv2} extends PETR with temporal modeling and incorporate ego-motion in the position embedding.
\ourMethod{} conducts the attention in image space and local 3D space to eliminate the variances in view transformation. \ourMethod{} could preserve the pros in single-view approaches and leverage the geometric information provided by multi-view images. 

\subsection{View Transformation}
The view transformation from the global view to local view in 3D scenes is an effective approach to boost performances for detection tasks. This could be treated as a normalization by aligning all the views, which could facilitate the learning procedure greatly.
Several LiDAR-based 3D detectors~\cite{shi2019pointrcnn, qi2018frustum, shi2020points, meyer2019lasernet, fan2021rangedet} estimate local coordinates rather than global coordinates for instances in the second stage, which could fully extract ROI features. 
For example, PointRCNN~\cite{shi2019pointrcnn} proposes the canonical 3D box refinement in the canonical system for more precise regression. 
To reduce the data variability for point cloud, AziNorm~\cite{chen2022azinorm} proposes a general normalization in the data pre-process stage.
Different from these methods, our method conduct view transformation for eliminating the extrinsic variances brought by multi cameras with camera-view position embedding.

%% file: sections/3_method.tex



\section{Our Approach}
We present a camera-view position embedding (CAPE) approach for multi-view 3D detection 
and construct the position embeddings
in each camera coordinate system.

\vspace{2mm}
\noindent\textbf{Architecture.}
We adopt the multi-view DETR framework,
a multi-view extension of DETR,
depicted in Figure~\ref{fig:overview}.
The input multi-view images,
$\{\mathbf{I}_1, \mathbf{I}_2,
\dots, \mathbf{I}_N\}$, 
are processed with the encoders to extract the image embeddings,
\begin{align}
\mathbf{X}_n = 
\operatorname{Encoder}(\mathbf{I}_n).
\end{align}
The $N$ image embeddings
are simply concatenated together,
$\mathbf{X}=[\mathbf{X}_1~\mathbf{X}_2~\dots~\mathbf{X}_N]$.
The associated position embeddings
are also concatenated together,
$\mathbf{P} = [\mathbf{P}_1~\mathbf{P}_2~\dots~\mathbf{P}_N]$. $\mathbf{X}_n, \mathbf{P}_n\in \mathbb{R}^{C\times I}$, where $I$ is the pixels number.

The decoder is similar to DETR decoder, with a stack of $L$ decoder layers
that is composed of
self-attention, cross-attention , and feed-forward network (FFN).
The $l$-th decoder layer is formulated as 
follows,
\begin{align}
\mathbf{O}_{l} = 
\operatorname{DecoderLayer}(\mathbf{O}_{l-1},
\mathbf{R},
\mathbf{X},
\mathbf{P}).
\end{align}
Here,
$\mathbf{O}_{l-1}$ and $\mathbf{O}_{l}$
are the output decoder embedding of the $(l-1)$th and the $l$th layer, respectively.
$\mathbf{R}$ are $3$D reference points following the design in \cite{zhu2020deformable,liu2022petr}.

Self-attention is the same as the normal DETR,
and 
takes the sum of $\mathbf{O}_{l-1}$
and the position embedding of the reference points $\mathbf{P}$ as input.
Our work lies in learning 3D position embeddings.
In contrast to PETR~\cite{liu2022petr} and PETRv2~\cite{liu2022petrv2}
that form the position embedding
in the global coordinate system,
we focus on learning
position embeddings
in the camera coordinate system
for cross-attention.


\begin{figure}[t]
\centering
\includegraphics[width=1\linewidth]{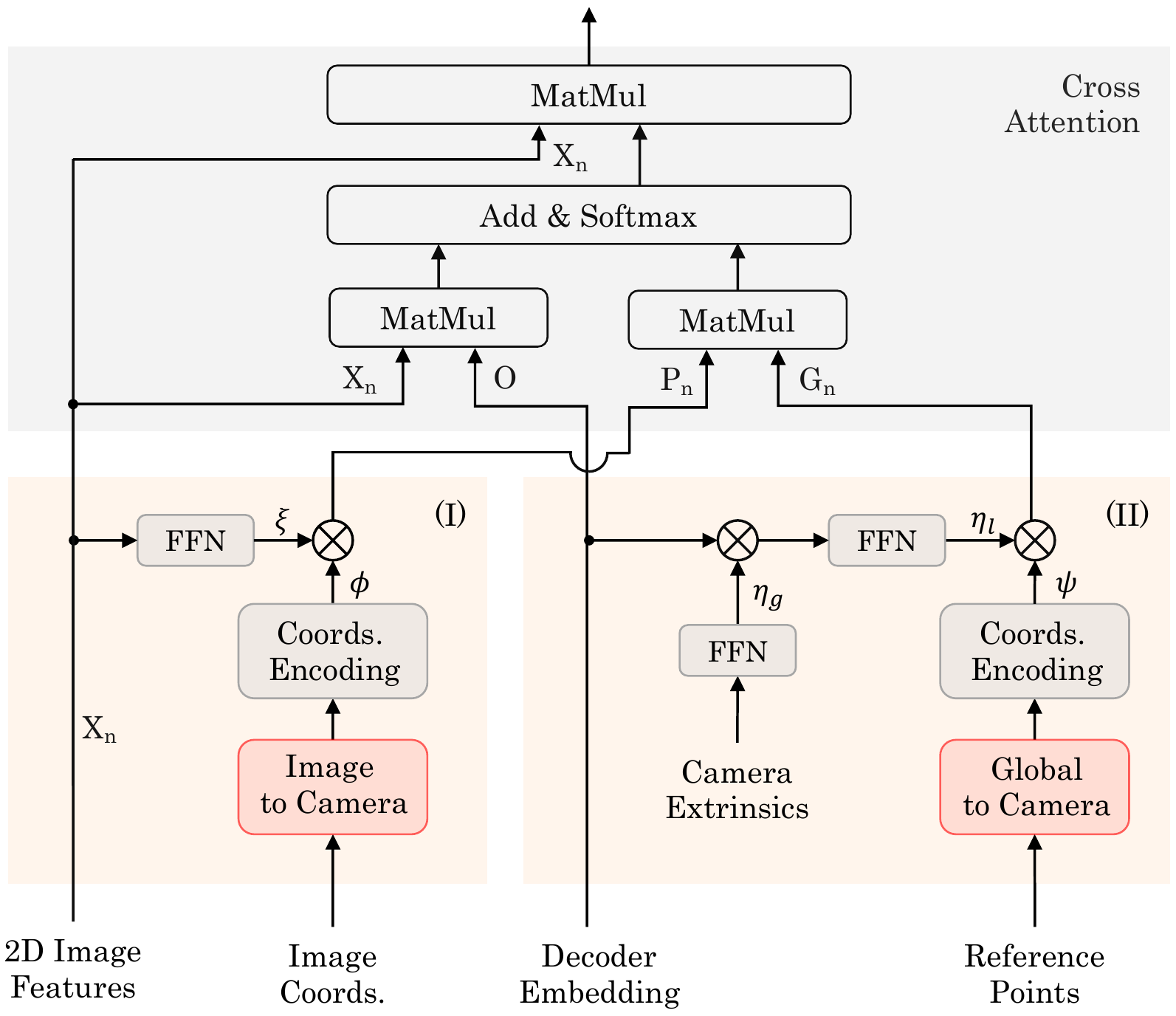}
\caption{The cross-attention module of \ourMethod{}. (I): Feature-guided Key Position Embedding (K-FPE); (II): Feature-guided Query Position Encoder (Q-FPE). The multi-head setting and fully connected layers are omitted for the sake of simplicity.}
\centering
\label{fig:transformer}
\vspace{-15pt}
\end{figure}

\vspace{2mm}
\noindent\textbf{Key Position Embedding Construction.}
We take one camera view (one image)
as an example
and describe how to construct the position embeddings
for one view.
Each $2$D position in the image plane
corresponds to $D$ $3$D coordinates along the predefined depth bins in the frustum:
$\{\mathbf{c}_1, \mathbf{c}_2,
\dots, \mathbf{c}_D\}$.
The $3$D coordinate in the image frustum is transformed 
to the camera coordinate system,
\begin{align}\mathbf{c}'_d = \mathbf{T}^{-1}_i\mathbf{c}_d,
\end{align}
where 
$\mathbf{T}_i$ is the intrinsic matrix for $i$-th camera.
The $D$ transformed $3$D coordinates 
$\{\mathbf{c}'_1, \mathbf{c}'_2,
\dots, \mathbf{c}'_D\}$
are mapped into the single embedding,
\begin{align}
    \mathbf{p}
    = \phi(\mathbf{c}'). 
    \label{eqn:KeyPE}
\end{align}
Here, $\mathbf{c}'$ is a $(D\times 3)$-dimensional vector, with 
$\mathbf{c}' = 
[{\mathbf{c}'}_1^\top~{\mathbf{c}'}_2^\top~\dots~{\mathbf{c}'}_D^\top]$.
$\phi$ is instantiated by a multi-layer perceptron (MLP) of two layers.

\vspace{2mm}
\noindent\textbf{Query Position Embedding Construction.}
We use a set of learnable $3$D reference points 
$\mathbf{R} = \{\mathbf{r}_1, \mathbf{r}_2,
\dots, \mathbf{r}_M\}$ in the global space 
to form the object queries. We transform the $M$ $3$D points
into the camera coordinate system for each view,
\begin{align}
    \bar{\mathbf{r}}_{nm}
    = \mathbf{T}_{n}^{e}\mathbf{r}_m,
\end{align}
where $\mathbf{T}_{n}^{e}$ is the extrinsic parameter matrix
for the $n$-th camera denoting the coordinate transformation from the global (LiDAR) system to the camera-view system.
The transformed $3$D coordinates
are then mapped into 
the query position embedding,
\begin{align}
    \mathbf{g}_{nm} = \psi(\bar{\mathbf{r}}_{nm}),
    \label{eqn:QPE}
\end{align}
where $\psi(\cdot)$ is a two-layer MLP.

Considering that
the decoder embedding $\mathbf{O}\in \mathbb{R}^{C\times M}$
and query position embeddings
are about different coordinate systems:
global coordinate system
and camera view coordinate system,
respectively,
we form the camera-dependent decoder queries
through concatenation(denote as $[\cdot]$):
\begin{align}
    \mathbf{Q}_n =[\mathbf{O}^\top~\mathbf{G}_n^\top]^\top,
\end{align}
where $\mathbf{G}_n = [\mathbf{g}_{n1} \mathbf{g}_{n2}\dots \mathbf{g}_{nM}]\in \mathbb{R}^{C\times M}$.
Accordingly,
the keys for cross-attention between queries and image features
are also formed through concatenation, 
$\mathbf{K}_n = [\mathbf{X}_{n}^\top~\mathbf{P}_{n}^\top]^\top$.
The $n-$view pre-normalized cross-attention weights $\mathbf{W}_n \in \mathbb{R}^{I\times M}$
is computed from:
\begin{align}
    \mathbf{W}_n = \mathbf{K}_n^\top \mathbf{Q}_n = \mathbf{X}_n^\top\mathbf{O} + \mathbf{P}_{n}^\top \mathbf{G}_{n}
\end{align}
The decoder embedding is updated by aggregating information from all views:
\begin{align}
\mathbf{O} \xleftarrow{}{} \sum_n \mathbf{X}_n\sigma(\mathbf{W}_n),
\end{align}
where $\sigma$ is the soft-max, note that projection layers are omitted for simplicity. More details can be seen in the appendix.
\noindent\textbf{Feature-guided Key and Query Position Embeddings.}
Similar to PETRv2~\cite{liu2022petrv2},
we make use of the image features to guide the key position embedding computation by learning the scaling weights
and update Eq.\ref{eqn:KeyPE} as:
\begin{align}
    \mathbf{p}
    = \phi(\mathbf{c}')\odot\xi(\mathbf{x}),
    \label{eqn:FGPC}
\end{align}
where $\xi(\cdot)$ is a two-layers MLP, $\odot$ denotes the element-wise multiplication and $\mathbf{x}$
is the image features at the corresponding position.
It is assumed to
provide some informative guidance
(e.g., depth). 

On the query side, inspired by conditional DETR~\cite{meng2021conditional}, we use the decoder embedding
to guide the query position embedding computation 
and update Eq.\ref{eqn:QPE} as:
\begin{align}
\mathbf{g}_{nm} = \psi(\bar{\mathbf{r}}_{nm})\odot \eta(\mathbf{o}_m, \mathbf{T}_{n}^{e}).
\end{align}
The extrinsic parameter matrix $\mathbf{T}_{n}^{e}$
is used to transform the spatial decoder embedding $\mathbf{o}_m$
to camera coordinate system,
for alignment with the reference point position embeddings.
Specifically, $\eta(\cdot)$ is instantiated as:
\begin{align}
\label{eq:cam_extrin_modulate}
\eta(\mathbf{o}_m, \mathbf{T}_{n}^{e}) = 
\eta_l(\mathbf{o}_m \odot \eta_g(\mathbf{T}_{n}^{e})).
\end{align}
Here $\eta_l(\cdot)$ and $\eta_g(\cdot)$ are two-layers MLP.




\begin{figure}[t]
\centering
\includegraphics[width=1\linewidth]{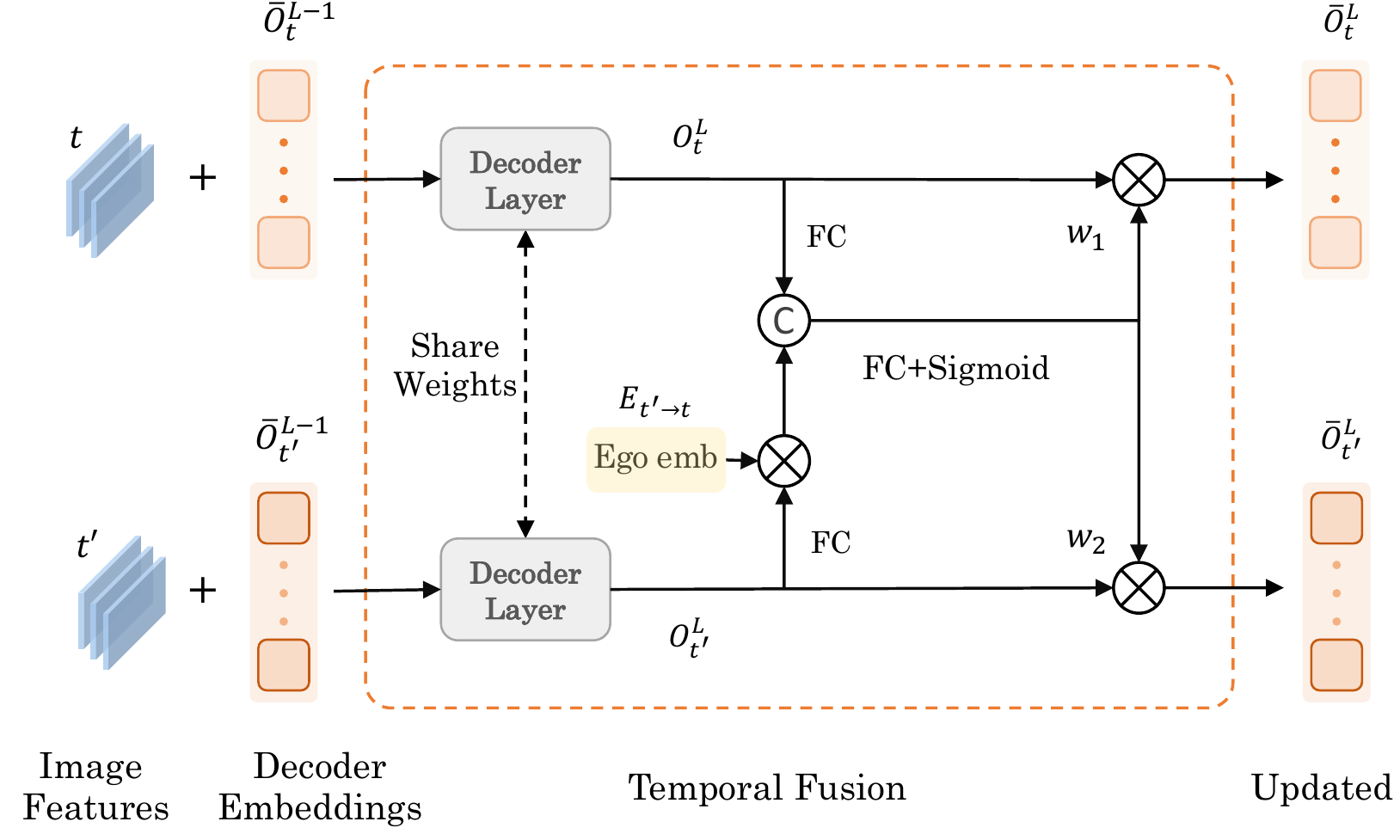}
\caption{The pipeline of our  temporal modeling. The ego-motion embedding modulates the decoder embedding for spatial alignment. Then each decoder embedding is scaled by the attention weights $(w1, w2)$ that predicted from the aligned embeddings.
}
\centering
\label{fig:temporal}
\end{figure}

\noindent\textbf{Temporal modeling
with ego-motion embedding.}
We utilize the previous frame information
to boost the detection
for the current frame.
The reference points of the previous frame
are transformed 
from the reference points
of the current frame
using the ego-motion matrix $\mathbf{M}$,
\begin{align}
    \mathbf{R}_{t'}
    = \mathbf{M}\mathbf{R}_t.
\end{align}
Considering the same moving objects may have different 3D positions in two frames, we build the separated decoder embeddings $(\mathbf{O}_{t}, \mathbf{O}_{t'})$ that can represent different position information for each frame. The interaction between the decoder embeddings of two frames for $l$-th decoder layer is formulated as follows:
\begin{align}
    (\bar{\mathbf{O}}_{t}^L,
    \bar{\mathbf{O}}_{t'}^L)
    = f(\mathbf{O}_{t}^{L}, \mathbf{O}_{t'}^{L}, \mathbf{M}).
\end{align}


We elaborate on the interaction between queries in two frames in Figure~\ref{fig:temporal}. Given that $\mathbf{O}_{t}$ and $\mathbf{O}_{t'}$ are not in the same ego coordinate system, we inject the ego-motion information into the decoder embedding $\mathbf{O}_{t'}$ for spatial alignment. Then we update decoder embeddings with channel attention weights generated from the concatenation of the decoder embeddings of two frames. Compared with using one set of queries learning objects in different frames, queries in our method have a stronger capability in positional learning.

\noindent\textbf{Heads and Losses.} The detection heads consist of the classification branch that predicts the probability of object classes and the regression branch that regresses 3D bounding boxes. The regression branch predicts the relative offsets \textit{w.r.t.} the coordinates of 3D reference points in the global system. As for the loss function, we adopt focal loss~\cite{lin2017focal} for classification $L_{cls}$ and L1 loss for regression $L_{reg}$ following prior works~\cite{wang2022detr3d,liu2022petr}. The label assignment strategy here is the Hungarian algorithm~\cite{kuhn1955hungarian}. Suppose that $\sigma$ is the assignment function, the loss for 3D object detection for the model without temporal modeling can be summarized as:
\begin{equation}
\begin{gathered}
L_{cur}(\mathbf{y}, \mathbf{\hat{y}}) = \lambda_{cls} L_{cls}(\mathbf{c}, \sigma{(\mathbf{\hat{c})}}) + L_{reg}(\mathbf{b}, \sigma{(\mathbf{\hat{b}})}),
\end{gathered}
\end{equation}
where $\mathbf{y}=(\mathbf{c}, \mathbf{b})$ and $\hat{\mathbf{y}}=(\mathbf{\hat{c}}, \mathbf{\hat{b}})$ denote the set of ground truths and predictions respectively. $\lambda_{cls}$ is a hyper-parameter to balance losses.
As for the network with temporal modeling, different from other methods supervise predictions only on the current frame, we predict results and supervise them on previous frames as auxiliary losses to enhance the temporal consistency. We use the center location and velocity of the ground truth on the current frame to generate the ground truths on the previous frame.
\begin{equation}
\begin{gathered}
L_{all} = L_{cur} + \lambda L_{prev}.
\end{gathered}
\end{equation}
$L_{cur}$ and $L_{prev}$ denote the losses for the current and previous frame separately. $\lambda$ is a hyper-parameter to balance losses.

%% file: sections/4_experiments.tex
\section{Experiments}
\subsection{Dataset}
We evaluate \ourMethod{} on the large-scale  nuScenes~\cite{caesar2020nuscenes} dataset. This dataset is composed of 1000 scene videos, with 700/150/150 scenes for training, validation, and testing set, respectively. Each sample consists of RGB images from 6 cameras and has 360 ° horizontal FOV. There are 20s video frames for each scene and 3D annotations are provided with every 0.5s. We report nuScenes Detection Score (NDS), mean Average Precision (mAP), and five True Positive (TP) metrics: mean Average Translation Error (mATE), mean Average Scale Error (mASE), mean Average Orientation Error (mAOE), mean Average Velocity Error (mAVE), mean Average Attribute Error (mAAE).

\subsection{Implementation Details}
We follow the PETR~\cite{liu2022petr} to report the results. We stack six transformer layers and adopt eight heads in the multi-head attention. 
Following other methods\cite{liu2022petr,wang2022detr3d, huang2021bevdet}, \ourMethod{} is trained with the pre-trained model FCOS3D~\cite{wang2021fcos3d} on validation dataset and with DD3D~\cite{park2021pseudo} pre-trained model on the test dataset as initialization. We use regular cropping, resizing, and flipping as data augmentations. The total batch size is eight, with one sample per GPU. 
We set $\lambda$ = 0.1 to balance the loss weight between the current frame and the previous frame and set $\lambda_{cls}$ = 2.0 to balance the loss weight between classification and regression. 
For validation dataset setting, we train \ourMethod{} for 24 epochs on 8 A100 GPUs with a starting learning rate of 2$e^{-4}$ that decayed with cosine annealing policy. 
For test dataset setting, we adopt denoise~\cite{zhang2022dino} for faster convergence. We train 24 epochs with CBGS on the single-frame setting. Then we load the single-frame model of CAPE as the pre-trained model for multi-frame training of CAPE-T and train 60 epochs without CBGS. 

\begin{table*}[t]
\small
\setlength{\tabcolsep}{3mm}
\centering
\caption{Comparison of recent works on the nuScenes \emph{test} set. $\ddagger$ is test time augmentation.
Setting ``S'':  only using single-frame information, ``M'': using multi-frame (two) information. 
``L'': using extra LiDAR data source as depth supervision.}
\vspace{-10pt}
\label{tab:main_result}
\resizebox{1\linewidth}{!}{
\begin{tabular}{l | c | c | c| c  c | c c c c c} 
 \hline
 Methods & Year & Backbone & Setting &  NDS$\uparrow$&mAP$\uparrow$&mATE$\downarrow$&mASE$\downarrow$&mAOE$\downarrow$&mAVE$\downarrow$& mAAE$\downarrow$ \\ [0.5ex] 
 \hline
BEVDepth$^\ddagger$~\cite{li2022bevdepth} & arXiv2022 & V2-99 & M, L & 0.600 & 0.503 & 0.445 & 0.245 & 0.378 & 0.320 & 0.126 \\
BEVStereo$^\ddagger$~\cite{li2022bevstereo} & arXiv2022 & V2-99 & M, L & 0.610 & 0.525 & 0.431 & 0.246 & 0.358 & 0.357 & 0.138 \\
 \hline
FCOS3D$^\ddagger$~\cite{wang2021fcos3d}& ICCV2021 &  Res-101 & S & 0.428 & 0.358 & 0.690 & 0.249 & 0.452 & 1.434 & 0.124  \\ 
PGD$^\ddagger$~\cite{wang2022probabilistic} & CoRL2022 & Res-101 & S & 0.448 & 0.386 & 0.626 & 0.245 & 0.451 & 1.509 & 0.127  \\ 
DETR3D~\cite{wang2022detr3d} & CoRL2022 & Res-101 & S & 0.479 & 0.412 & 0.641 & 0.255 & 0.394 & 0.845 & 0.133  \\ 
BEVDet$^\ddagger$~\cite{huang2021bevdet} & arXiv2022 & V2-99   & S & 0.488 & 0.424 & 0.524 & 0.242 & 0.373 & 0.950 & 0.148  \\ 
PETR~\cite{liu2022petr} & ECCV2022 & V2-99 & S & 0.504 & 0.441 & 0.593 & 0.249 & 0.383 & 0.808 & 0.132  \\ 
\cellcolor{Gray}\ourMethod{} & \cellcolor{Gray} - & \cellcolor{Gray}V2-99  & \cellcolor{Gray}S & 
\cellcolor{Gray}\textbf{0.520} & 
\cellcolor{Gray}\textbf{0.458} & 
\cellcolor{Gray}0.561 & 
\cellcolor{Gray}0.252 &
\cellcolor{Gray}0.389 & 
\cellcolor{Gray}0.758 &
\cellcolor{Gray}0.132  \\ 
 \hline
BEVFormer~\cite{li2022bevformer}& ECCV2022 & V2-99 & M & 0.569 & 0.481 & 0.582 & 0.256 & 0.375 & 0.378 & 0.126  \\
BEVDet4D$^\ddagger$~\cite{2022BEVDet4D} & arXiv2022 & Swin-B & M & 0.569 & 0.451 & 0.511 & 0.241 & 0.386 & 0.301 & 0.121  \\
PETRv2~\cite{liu2022petrv2} & arXiv2022 & V2-99  & M & 0.582 & 0.490 & 0.561 & 0.243 & 0.361 & 0.343 & 0.120  \\
\cellcolor{Gray}\ourMethod{}-T & \cellcolor{Gray} - & \cellcolor{Gray}V2-99 & \cellcolor{Gray}M & 
\cellcolor{Gray}\textbf{0.610} & 
\cellcolor{Gray}\textbf{0.525} & 
\cellcolor{Gray}0.503 & 
\cellcolor{Gray}0.242 & 
\cellcolor{Gray}0.361 & 
\cellcolor{Gray}0.306 & 
\cellcolor{Gray}0.114 \\ 
 \hline
\end{tabular}
}
\vspace{-0.1cm}
\end{table*}

\begin{table*}[t]
\small
\setlength{\tabcolsep}{3mm}
\centering
\caption{Comparison on the nuScenes \emph{validation} set with large backbones. All listed methods are trained with 24 epochs without CBGS. }
\vspace{-10pt}
\label{tab:main_result_val}
\resizebox{1\linewidth}{!}{
\begin{tabular}{l | c | c | c | c c | c c c c c} 
 \hline
 Methods & Backbone &  Resolution & Setting & NDS$\uparrow$&mAP$\uparrow$&mATE$\downarrow$&mASE$\downarrow$&mAOE$\downarrow$&mAVE$\downarrow$& mAAE$\downarrow$ \\ [0.5ex] 
 \hline
BEVDet~\cite{huang2021bevdet} & Swin-B & $1408 \times 512$  & S & 0.417 & 0.349 & 0.637 & 0.269 & 0.490 & 0.914 & 0.268  \\
PETR~\cite{liu2022petr}   & V2-99   & $1600 \times 900$ & S & 0.455 & 0.406 & 0.736 & 0.271 & 0.432 & 0.825 & 0.204  \\
\cellcolor{Gray}\ourMethod{}& \cellcolor{Gray}V2-99 & \cellcolor{Gray}$1600 \times 900$ & 
\cellcolor{Gray}S &
\cellcolor{Gray}\textbf{0.479} &
\cellcolor{Gray}\textbf{0.439} & 
\cellcolor{Gray}0.683 & 
\cellcolor{Gray}0.267 & 
\cellcolor{Gray}0.427 & 
\cellcolor{Gray}0.814 & 
\cellcolor{Gray}0.197  \\
 \hline
BEVDet4D~\cite{2022BEVDet4D} & Swin-B & $1600 \times 900$ & M & 0.515 & 0.396& 0.619 & 0.260 & 0.361 & 0.399 & 0.189 \\
PETRv2~\cite{liu2022petrv2} & V2-99 & $800 \times 320$    & M & 0.503 & 0.410 & 0.723 & 0.269 & 0.453 & 0.389 & 0.193 \\
\cellcolor{Gray}\ourMethod{}-T& \cellcolor{Gray}V2-99 & \cellcolor{Gray}$800 \times 320$ & 
\cellcolor{Gray}M &
\cellcolor{Gray}\textbf{0.536} & 
\cellcolor{Gray}\textbf{0.440} & 
\cellcolor{Gray}0.675 &
\cellcolor{Gray}0.267 &
\cellcolor{Gray}0.396 &
\cellcolor{Gray}0.323 &
\cellcolor{Gray}0.185 \\
 
 \hline
\end{tabular}
}
\vspace{-0.2cm}
\end{table*}

\begin{table}[!t]
\small
\setlength{\tabcolsep}{3mm}
\caption{Comparison on the nuScenes \emph{validation} set with ResNet backbone. “†”: the results are reproduced for a fair comparison with our method(trained with 24 epochs). 
}
\vspace{-10pt}
\label{tab:res50_101}
\resizebox{\linewidth}{!}{
\begin{tabular}{l|c c c| c c} 
 \hline
 Method & Backbone & Resolution & CBGS &  NDS$\uparrow$&mAP$\uparrow$ \\ [0.5ex] 
 \hline
PETR†~\cite{liu2022petr} & Res-50 &  $1408 \times 512$ & \XSolidBrush & 0.367 & 0.317 \\
\cellcolor{Gray}CAPE & \cellcolor{Gray}Res-50 & \cellcolor{Gray}$1408 \times 512$ & \cellcolor{Gray}\XSolidBrush & \cellcolor{Gray}\textbf{0.380} & \cellcolor{Gray}\textbf{0.337} \\
\hline
FCOS3D~\cite{wang2021fcos3d} & Res-101 & $1600 \times 900$ & \Checkmark & 0.415 & 0.343 \\ 
PGD~\cite{wang2022probabilistic} & Res-101 & $1600 \times 900$ & \Checkmark  & 0.428 & 0.369 \\ 
DETR3D~\cite{wang2022detr3d}& Res-101 & $1600 \times 900$ & \Checkmark & 0.434 & 0.349 \\ 
BEVDet~\cite{huang2021bevdet} & Res-101 & $1056 \times 384$ & \Checkmark & 0.396 & 0.330 \\
PETR~\cite{liu2022petr} & Res-101 & $1600 \times 900 $ & \Checkmark & 0.442 & 0.370\\
\cellcolor{Gray}CAPE & \cellcolor{Gray}Res-101 & \cellcolor{Gray}$1600 \times 900 $ &  \cellcolor{Gray}\Checkmark & \cellcolor{Gray}\textbf{0.463} & \cellcolor{Gray}\textbf{0.388} \\ 
 \hline
BEVFormer~\cite{li2022bevformer} & Res-50 & $800 \times 450$& \XSolidBrush & 0.354 & 0.252\\ 
PETRv2†~\cite{liu2022petrv2} & Res-50 &  $704 \times 256$ & \XSolidBrush & 0.402 & 0.293\\ 
\cellcolor{Gray}CAPE-T & \cellcolor{Gray}Res-50 &  \cellcolor{Gray}$704 \times 256$ &\cellcolor{Gray}\XSolidBrush & \cellcolor{Gray}\textbf{0.442} & \cellcolor{Gray}\textbf{0.318} \\ 
\hline
BEVFormer~\cite{li2022bevformer} & Res-101 & $1600 \times 640$& \Checkmark & 0.517 & 0.416\\ 
PETRv2~\cite{liu2022petrv2} & Res-101 & $1600 \times 640$& \Checkmark & 0.524 & 0.421 \\ 
\cellcolor{Gray}CAPE-T & \cellcolor{Gray}Res-101 &  \cellcolor{Gray}$1600 \times 640$ & \cellcolor{Gray}\Checkmark & \cellcolor{Gray}\textbf{0.533} & \cellcolor{Gray}\textbf{0.431}\\ 
 \hline
\end{tabular}
}
\vspace{-0.2cm}
\end{table}

\subsection{Comparison with State-of-the-art}
We show the performance comparison in the nuScenes test set in Tab.~\ref{tab:main_result}. We first compare the \ourMethod{} with state-of-the-art methods on the single-frame setting and then compare \ourMethod{}-T(the temporal version of \ourMethod{}) with methods that leverage temporal information. 
As for the model on the single-frame setting, as far as we know, \ourMethod{}(NDS=52.0) could achieve \textbf{the first place} on nuScenes benchmark compared with vision-based methods with the single-frame setting.
As for the model with temporal modeling, \ourMethod{}-T still outperforms all listed methods. \ourMethod{} achieves \highNDS{} on NDS and \highmAP{} on mAP. We point out that using LiDAR as supervision could largely improve the mATE metric, thus it's not fair to compare methods (w/wo LiDAR supervision) together. Nevertheless, even compared with contemporary methods leveraging LiDAR as supervision, \ourMethod{} outperforms BEVDepth~\cite{li2022bevdepth} $1.0\%$ on NDS and $2.2\%$ on mAP and achieves comparable results to BEVStereo~\cite{li2022bevstereo}. 

We further show the performance comparison on the nuScenes validation set in Tab.~\ref{tab:main_result_val} and Tab.~\ref{tab:res50_101}. It could be seen that \ourMethod{} surpasses our baseline to a large margin and performs well compared with other methods. 

\subsection{Ablation Studies} In this section, we validate the effectiveness of our designed components in \ourMethod{}.
 We use the $1600 \times 900$ resolution for all single-frame experiments and the $800 \times 320$ resolution for all multi-frames experiments.

\vspace{2mm}
\noindent\textbf{Effectiveness of camera view position embedding.} We validate the effectiveness of our proposed camera view position embedding in Tab~\ref{tab:ablation_view-invariant_decouple}. In Setting(a), we simply adopt PETR with feature-guided position embedding as our baseline. When we adopt the bilateral attention mechanism with 3D position embedding in the LiDAR system in Setting(b), the performance can be improved by $0.5\%$ on NDS. When using camera 3D position embedding without bilateral attention mechanism in Setting(c), the model could not converge and only get $2.5\%$ NDS. It indicates that the 3D position embedding in the camera system should be decoupled with output queries in the LiDAR system. The best performances could be achieved when we use camera view position embeddings along with the bilateral attention mechanism.
For fair comparison with 3D position embedding in the LiDAR system, camera view position embedding improves $1.4\%$ in NDS and $2.4\%$ in mAP (see Setting(b) and (d)).

\begin{table}[!t]
\small
\centering
\setlength{\tabcolsep}{1.5mm}
\caption{Ablation study of camera view position embedding and bilateral attention mechanism on the nuScenes validation set.
``Cam.View'': 3D position embeddings are constructed in the camera system, otherwise 3D position embeddings are constructed in the global (LiDAR) system.``Bilateral'': bilateral attention mechanism is adopted in the decoder.}
\label{tab:ablation_view-invariant_decouple}
\vspace{-10pt}
\resizebox{\linewidth}{!}{
\begin{tabular}{c| c c| c c c c} 
 \hline
 Setting & Cam.View & Bilateral & NDS$\uparrow$&mAP$\uparrow$&mATE$\downarrow$&mAOE$\downarrow$ \\ [0.5ex] 
 \hline
(a) & & & 0.460 & 0.409 & 0.734 & 0.424  \\ 
(b) & & $\checkmark$ & 0.465 & 0.415 & 0.708 & 0.420 \\ 
 (c) & $\checkmark$ &  & 0.025 & 0.100 & 1.117 & 0.918 \\
 (d) & $\checkmark$ & $\checkmark$ &  0.479 & 0.439 & 0.683 & 0.427 \\
 \hline
\end{tabular}
}
\vspace{-10pt}
\end{table}
\begin{table}[t]
\small
\setlength{\tabcolsep}{2.3mm}
\centering
\caption{Ablation study of feature guided position embedding in queries and keys on the nuScenes validation set. ``Q-FPE'': using feature-guided design in queries.``K-FPE'': using feature-guided design in keys. }
\label{tab:ablation_KPFE_and_QFPE}
\vspace{-10pt}
\resizebox{\linewidth}{!}{
\begin{tabular}{c | c c | c c c c} 
 \hline
 Setting & Q-FPE & K-FPE & NDS$\uparrow$&mAP$\uparrow$&mATE$\downarrow$&mAOE$\downarrow$ \\ [0.5ex] 
 \hline
(a) & & & 0.447 & 0.415 & 0.719 & 0.515 \\ 
(b) &$\checkmark$ & & 0.449 & 0.421 & 0.693 & 0.551 \\ 
(c) & &$\checkmark$ & 0.463 & 0.420 & 0.700 & 0.473 \\ 
(d) &$\checkmark$ &$\checkmark$ & 0.479 & 0.439 & 0.683 & 0.427 \\
 \hline
\end{tabular}
}
\vspace{-15pt}
\end{table}

\vspace{2mm}
\noindent\textbf{Effectiveness of feature-guided position embedding.}
Tab.~\ref{tab:ablation_KPFE_and_QFPE} shows the effect of feature-guided position embedding in both queries and keys. 
Different from the FPE in PETRv2\cite{paszke2019pytorch}, our K-FPE here is formed under the local camera-view coordinate system instead of the global coordinate system. Compared with Setting(a) and (b), we find the Q-FPE increases the location accuracy (see the improvement of mAP and mATE), but decreases the orientation performance in mAOE, mainly owing to the lack of image appearance information in the local view attention. 
Compared with Setting(a) and (c), using K-FPE could improve $1.6\%$ on NDS and $4.2\%$ on mAOE, which benifits from more precise depth and orientation information in image appearance features. Compared with Setting(c) and (d), adding Q-FPE could bring $1.6\%$ and $1.9\%$ gain on NDS and mAP further. The benefit brought by Q-FPE could be explained by 3D anchor points being refined by input queries in high-dimensional embedding space. It could see that using both Q-FPE and K-FPE achieves the best performances. 

\vspace{2mm}
\noindent\textbf{Effectiveness of the temporal modeling approach.} We show the necessity of using a set of queries for each frame in Tab.~\ref{tab:ablation_group_queries}. It could be observed that decomposing queries into different frames could improve $0.9\%$ on NDS and $0.8\%$ on mAP. With the multi-group design, one object query could correspond to one instance on each frame respectively, which 
is proper for DETR-based paradigms. Similar conclusion is also observed in 2D instance segmentation tasks~\cite{wu2021seqformer}. Since we use multi groups of queries, auxiliary supervision on previous frames can be added to better align object queries between frames. Meanwhile, the generated ground truth on the previous frame could be treated as a type of data augmentation to avoid overfitting. When we 
adopt the previous loss, the mAP increases $0.5\%$, which proves the validity of supervision on multi-frames. Compared with Setting(a) and (c), our temporal modeling approach could improve $1.4\%$ on NDS and $1.2\%$ on mAP on the validation dataset. 

\begin{table}[t]
\small
\setlength{\tabcolsep}{2.3mm}
\centering
\caption{Ablation studies of our temporal modeling approach with ego-motion embedding on the nuScenes validation set. We validate the necessity of using a group of queries for each frame and the effectiveness of conducting the supervision on previous frames. ``QT'': whether sharing \textbf{Q}ueries for each frame in \textbf{T}emporal modeling. ``LP'': using auxiliary \textbf{L}oss on \textbf{P}revious frames. 
}
\label{tab:ablation_group_queries}
\vspace{-10pt}
\resizebox{\linewidth}{!}{
\begin{tabular}{c|c c | c c c c c} 
 \hline
 Setting & QT & LP & NDS$\uparrow$&mAP$\uparrow$&mAOE$\downarrow$&mAVE$\downarrow$ \\ [0.5ex] 
 \hline
 (a) & Share &  & 0.522 & 0.428 & 0.454 & 0.341 \\ 
 (b) & Not Share & & 0.531 & 0.436 & 0.401 & 0.346 \\
 (c) & Not Share & $\checkmark$ & 0.536 & 0.440 & 0.396 & 0.323 \\
 \hline
\end{tabular}
}
\vspace{-0.2cm}
\end{table}

\begin{table}[t]
\small
\setlength{\tabcolsep}{1.7mm}
\centering
\caption{Ablation study of our temporal fusion approach on nuScenes validation set. ``Ego'': using ego motion embedding in the fusion process; ``concat + ML'': simply concatenating queries in two frames and using a two-layer MLP to output fused queries; ``channel att'': learning channel attention weights for each query.}
\label{tab:ablation_fusion_methods}
\vspace{-10pt}
\resizebox{1\linewidth}{!}{
\begin{tabular}{c|c c| c c c c} 
 \hline
 Setting & Fusion ways & Ego & NDS$\uparrow$&mAP$\uparrow$&mAOE$\downarrow$&mAVE$\downarrow$ \\ [0.5ex] 
 \hline
 (a) & concat + MLP & &  0.527 & 0.432 & 0.413 & 0.343 \\ 
 (b) & concat + MLP & $\checkmark$ & 0.531 & 0.439 & 0.432 & 0.318 \\
 (c) & channel att & & 0.530 & 0.432 & 0.400 & 0.333 \\
 (d) & channel att & $\checkmark$ & 0.536 & 0.440 & 0.396 & 0.323 \\
 
 \hline
\end{tabular}
}
\vspace{-15pt}
\end{table}


\begin{figure*}[t]
\centering
\includegraphics[width=1.0\linewidth]{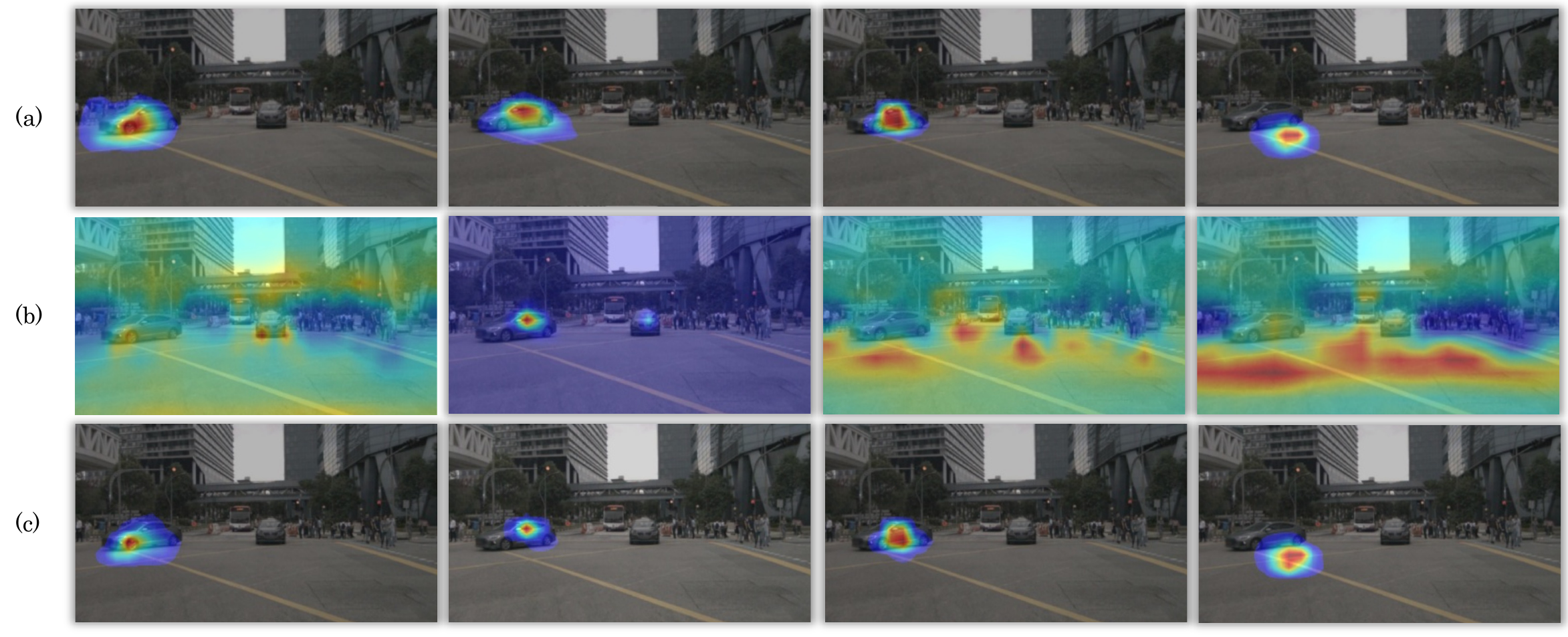}
\vspace{-10pt}
\caption{Visualization of attention maps from an object query in the last decoder layer. Four heads out of eight are shown here. We only show a single view for simplicity, (a): the normalized $\mathbf{G}_{n}^\top\mathbf{P}_{n}$ (local view attention maps), (b): the normalized $\mathbf{X}_n^\top\mathbf{O}$ (global view attention maps), (c): the overall attention maps that are the normalized weights of the summation of the former two items.
Note that we only visualize the attention weights that are greater than $1e^{-4}$ for better visualization.
}
\centering
\label{fig:vis_attn}
\vspace{-10pt}
\end{figure*}

\vspace{2mm}
\noindent\textbf{Effectiveness of different fusion approaches.}
The fusion module is used to fuse different frames for temporal modeling. We explore some common fusion approaches for temporal fusion in Tab.~\ref{tab:ablation_fusion_methods}. We first try a simple fusion approach ``concat with MLP'', and achieve $52.7\%$ on NDS, which has $0.5\%$ improvement compared with sharing queries. Considering queries on each frame have similar semantic information and different positional information, we propose the fusion model inspired by the channel attention. As is seen in Tab.~\ref{tab:ablation_fusion_methods}, our proposed fusion approach ``channel att'' achieves higher performance compared to simple concatenation operation. 
We claim that the performance gain is not from the increased parameters since only three fully-connected layers are added in our model.
Since queries are defined in each frame's system and the ego motion occurs between frames, we encode the ego-motion matrix as a high dimensional embedding to align queries in the current frame's system. With ego-motion embeddings, our fusion approach could further improve $0.6\%$ on NDS and $0.8\%$ on mAP.  

\subsection{Visualization} 
We show the visualization of attention maps in Figure ~\ref{fig:vis_attn} from 4 heads out of 8 heads. We display attention maps after the soft-max normalized operation. From the up-bottom way in each row, there are local view attention maps, global view attention maps, and overall attention maps separately. We obverse and draw three conclusions from visualization results. Firstly, local view attention mainly tends to highlight the neighbor of objects, such as the front, middle, and bottom of the objects, while global view attention pays more attention to the whole of images, especially on the ground plane and the same type of objects, as shown in  Figure ~\ref{fig:vis_attn} (a). 
This phenomenon indicates that local view attention and global view attention are complementary to each other. Secondly, as shown in Figure ~\ref{fig:vis_attn} (a) and (c), it can be seen that overall attention maps are highly similar to local view attention maps, which means the local view attention maps play the dominant role compared to global view attention. Thirdly, we could obverse that the overall attention maps are further concentrating on the foreground objects in a fine-grained way, which implies superior localization accuracy.

\begin{figure}[t]
\centering
\includegraphics[width=0.9\linewidth]{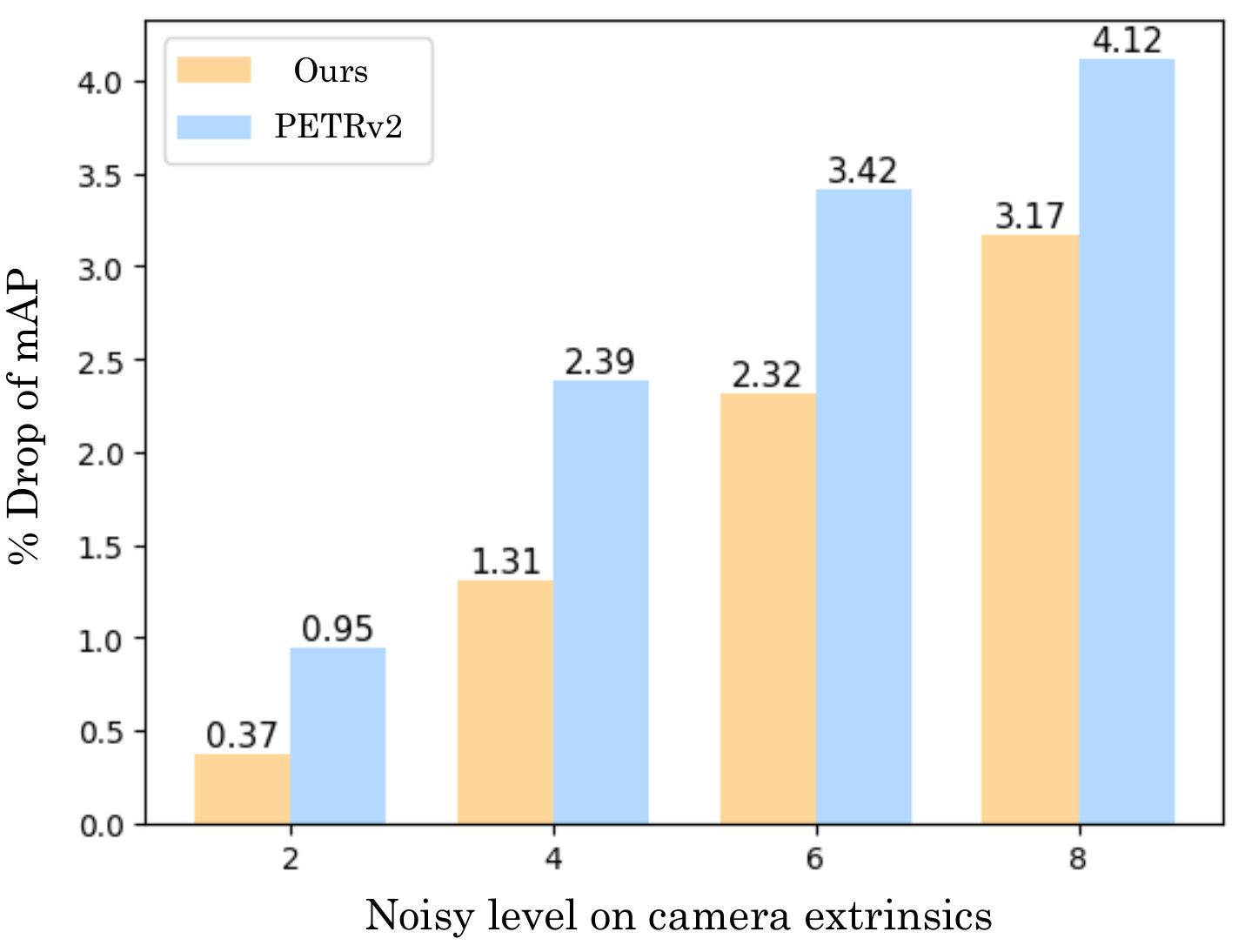}
\vspace{-10pt}
\caption{The performance drop of PETRv2 and \ourMethod{}-T under different camera extrinsics noise levels on nuScenes validation set.}
\centering
\label{fig:robustness}
\vspace{-10pt}
\end{figure}

\subsection{Robustness Analysis}
We evaluate the robustness of our method on camera extrinsic interference in this section. Camera extrinsic interference is an unavoidable dilemma caused by calibration errors, vehicle jittering, etc. We imitate extrinsics noises on the rotation with different noisy levels following PETRv2~\cite{liu2022petrv2} for a fair comparison. Specifically, we randomly sample an angle within the specific range and then multiply the generated noisy rotation matrix by the camera extrinsics in the inference. We present the performance drop of metric mAP on both PETRv2 and \ourMethod{}-T in Fig.\ref{fig:robustness}. We could see that our method has more robust performances on all noisy levels compared to PETRv2 ~\cite{liu2022petrv2} when facing extrinsics interference. For example, in the noisy level setting $R_{max} = 4$, \ourMethod{}-T drops 1.31\% while PETRv2 drops 2.39\%, which shows the superiority of camera-view position embeddings. 



%% file: sections/5_conclusion.tex
\section{Conclusion}
\label{conclusion}
In this paper, we study the 3D positional embeddings of sparse query-based approaches for multi-view 3D object detection and propose a simple yet effective method CAPE. We form the 3D position embedding under the local camera-view system rather than the global coordinate system, which largely reduces the difficulty of the view transformation learning. Furthermore, we extend our CAPE to temporal modeling by exploiting the fusion between separated queries for temporal frames. It achieves state-of-the-art performance even without LiDAR supervision, and provides a new insight of position embedding in multi-view 3D object detection.

\noindent\textbf{Limitation and future work.}  The computation and memory cost would be unaffordable when it involves the temporal fusion of long-term frames. In the future, we will dig deeper into more efficient spatial and temporal interaction of 2D and 3D features for autonomous driving systems.